\documentclass{article}

\usepackage[preprint]{neurips_2026}

\usepackage[utf8]{inputenc} 
\usepackage[T1]{fontenc}    
\usepackage{hyperref}       
\usepackage{url}            
\usepackage{booktabs}       
\usepackage{amsfonts}       
\usepackage{nicefrac}       
\usepackage{microtype}      
\usepackage{xcolor}         

\usepackage{graphicx} 
\usepackage{subcaption} 
\usepackage{algorithm}
\usepackage{algorithmicx}
\usepackage{algpseudocode}
\usepackage{booktabs}
\usepackage{enumitem}
\usepackage{colortbl}
\usepackage{caption}
\usepackage{amsmath}
\usepackage{amssymb}
\usepackage{tabularx}
\usepackage{multirow}
\usepackage{makecell}
\usepackage{wrapfig}

\definecolor{mygreen}{RGB}{0,160,0}

\definecolor{myred}{RGB}{178,34,34}

\title{PraMem: Practice-derived Experiential Memory for Long-horizon Behavior Prediction}

\author{
Zhuoqun Li${}^{1,2}$,
Boxi Cao${}^{1}$,
Jiawei Chen${}^{1,2}$,
Hanshu Zhou${}^{3}$,
Ruoxi Xu${}^{1,2}$,
\\
{\bf Guiping Jiang${}^{4}$,}
{\bf Ruotong Pan${}^{4}$,}
{\bf Tingting Gao${}^{4}$,}
{\bf Han Li${}^{4}$,}
{\bf Xiangyu Wu${}^{4}$,}
\\
{\bf Hongyu Lin${}^{1}$,}
{\bf Yaojie Lu${}^{1}$,}
{\bf Xianpei Han${}^{1}$,}
{\bf Le Sun${}^{1}$}
\\
${}^{1}$Chinese Information Processing Laboratory,
Institute of Software, Chinese Academy of Sciences\\
${}^{2}$University of Chinese Academy of Sciences \\
${}^{3}$Fudan University \\
${}^{4}$Kuaishou Technology \\
{\tt \{lizhuoqun2021, caoboxi\}@iscas.ac.cn} 
}

\begin{document}

\maketitle

\begin{abstract}
Long-horizon behavior prediction aims to infer a user's next action based on a lengthy historical sequence, playing a crucial role in artificial intelligence field.
The rise of large language models (LLMs) offers a promising direction for sequential behavior prediction, yet LLMs struggle with latent behavioral pattern induction and model-intrinsic cognitive biases when tackling long-horizon behavior prediction.
Prior memory management methods follow a context-compression paradigm that attempts to address this task by alleviating the historical sequence burden, yet fail to resolve the core challenges.
In this paper, we advocate a paradigm shift that reframes the lengthy historical sequence from a burden into a valuable resource to be exploited, and accordingly propose PraMem, which conducts beforehand practice over the lengthy historical sequence to build an experiential memory, thereby serving as the assisted input for accurate long-horizon behavior prediction.
Extensive experiments across diverse tasks demonstrate that PraMem achieves superior performance than prior methods, and more in-depth analyses provide valuable insights into the mechanism and evolution of the experiential memory. Code: \textcolor{blue}{https://github.com/icip-cas/PraMem}. 
\end{abstract}

\section{Introduction}

Long-horizon behavior prediction aims to infer a user's action in the current scene by unbiasedly inducing behavioral patterns from a lengthy historical sequence of scene-action records, serving as a critical foundation for recommendation systems, cognitive science, and interactive artificial intelligence~\citep{mcclelland2009place,petrovic2018artificial,ren2019lifelong,pi2019practice}.
Recently, formulating sequential behavior prediction as an autoregressive generation task with large language models (LLMs) has emerged as a promising direction, but long-horizon sequences spanning thousands of steps still pose a substantial challenge~\citep{zhang2024generative,chen2026towards}.
On one hand, latent behavioral pattern induction from ultra-long sequences is highly difficult, as fine-grained factors influencing the next action (e.g. \textit{users' preferences for brand, price, quality, and appearance when selecting products}) are  hidden and scattered across the lengthy sequence, making them hard to be accurately captured~\citep{lin2024rella,liu2025learning}.
On the other hand, LLMs are subject to  model-intrinsic cognitive biases when performing this task, such as \textit{the tendency to follow majority, prejudice from training, and disproportionate attention to certain records in sequence}, all of which further degrade  performance on long-horizon behavior prediction~\citep{liu2024lost,weng2025do}.

To enhance LLMs on long-horizon behavior prediction tasks, previous studies have explored memory management to  alleviate the long-context burden through information compression, which removes redundant text of the historical sequence via extraction and aggregation, and leverages retrieval to retain only the most relevant behavioral records for prediction~\citep{zheng2024harnessing,wang2024recmind,xi2024memocrs,wang2025user}.
However, such a context-compression paradigm fails to effectively address the core challenges of long-horizon behavior prediction. 
First, information compression is not equivalent to the behavioral pattern induction. LLMs still need to induce latent behavioral patterns on the fly from fragmented behavioral records during the prediction stage, which makes the reliability of induced behavioral patterns hard to guarantee~\citep{li2025mirage,jiang2025know}.
Second, previous methods focus solely on memory management over the historical sequence, lacking any mechanism to perceive or correct model-intrinsic cognitive biases of the LLM. This means that even when the provided information is sufficiently accurate and relevant, intrinsic biases of the LLM can still lead to incorrect prediction ~\citep{weng2025do,sumita2025cognitive}.

Limitations of above context-compression paradigm stem from the guiding philosophy, which treats historical sequences mainly as a long-context burden to alleviate. In contrast, dialectical philosophy suggests that burden and resource are a unity of opposites, where the two can transform into each other under the right condition~\citep{hegel2014science,engels1960dialectics}.
In this task, while lengthy historical sequences impose a heavy burden, they also contain valuable resources for improving LLM performance, as each segment of the historical sequence naturally forms a practice sample for sequential behavior prediction with a ground-truth label.
Therefore, the dialectical philosophy can inspire a new paradigm, instead of passively alleviating the burden of  historical sequences through compression, LLMs can proactively leverage these valuable resources for  practice.
Through iterative trial-and-error, LLMs can induce latent behavioral patterns in the historical sequence and guarantee their reliability, as well as find out model-intrinsic biases for current user's behavior prediction and alert against them.

\begin{figure*}[t]
\centering
\includegraphics[width=\linewidth]{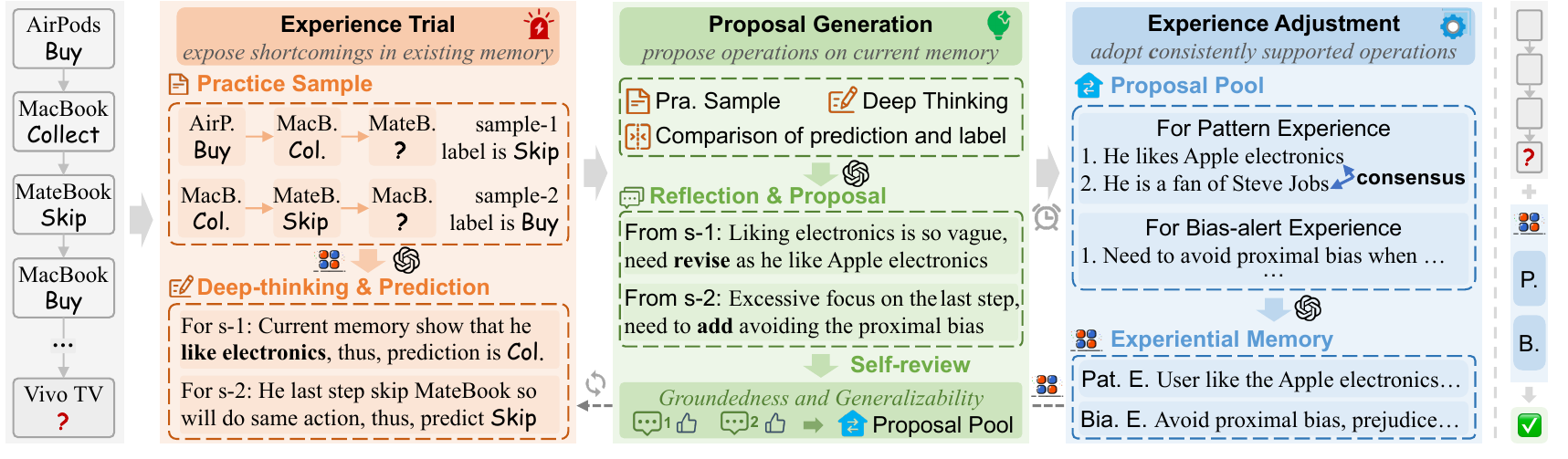} 
\caption{PraMem iteratively performs existing experience trial, reflective proposal generation, and consensual experience adjustment to dynamically maintain the pattern experience and bias-alert experience, and finally uses this experiential memory for accurate long-horizon behavior prediction.}
\label{fig:method}
\end{figure*}

Inspired by above ideas, we propose \textbf{PraMem}, a training-free framework that builds a time-evolving experiential memory by beforehand practice on the historical sequence, thereby assisting accurate prediction.
Specifically, as illustrated in Figure~\ref{fig:method}, the experiential memory consists of pattern experience for presenting user behavioral patterns and bias-alert experience for alerting LLM intrinsic biases.
In order to dynamically maintain the experiential memory, PraMem iteratively performs existing experience trial, reflective proposal generation, and consensual experience adjustment.
First, to expose shortcomings in existing experience, PraMem draws segments from the historical sequence to construct practice samples with ground-truth labels, and prompts the LLM to predict via explicit deep thinking under current experiential memory.
Second, to revise, prune, or supplement  current experiential memory, PraMem performs reflection on the practice sample, thinking process, and prediction-label comparison, generating a batch of reflective proposals, from which reliable ones are retained via a self-review mechanism and added to a proposal pool.
Finally, to prevent unstable adjustment of experiential memory caused by occasional behaviors, PraMem performs consensus-driven adjustment based on the proposal pool at fixed round intervals, where only operations consensually supported by multiple proposals are adopted, enabling stable evolution of experiential memory.

Within the above framework, the quality of reflective proposals is a key factor in ensuring the overall effectiveness, requiring each proposal to be both genuinely grounded in the practice sample and generalizable beyond the practice sample.
To this end, we design a self-review mechanism to filter out unreliable proposals by applying transformations to the practice sample. 
First, to identify proposals without sufficient groundedness with respect to the practice sample, we perturb the historical sequence of the sample so that its underlying behavioral patterns change significantly. If the original proposal still holds after the perturbation, it indicates that the proposal is not genuinely grounded in the practice sample. 
Second, to identify proposals that lack generalizability, we rewrite the prediction scene  of the practice sample to generate multiple similar but distinct virtual scenes, forming a multiple-choice task of scenes. If a proposal can unambiguously guide the correct selection of the true scene, it indicates that the proposal is overly specific to current practice sample rather than generally applicable.
Through the self-review mechanism, only reflective proposals that are both well-grounded and generalizable are added to the proposal pool, thereby ensuring the overall effectiveness of PraMem.

In experiments, we evaluate PraMem across various long-horizon behavior prediction tasks and compare it with several strong memory management baselines, the results demonstrate that PraMem achieves superior performance, confirming PraMem is an effective solution for long-horizon behavior prediction. 
Additionally, comprehensive ablation studies verify the effectiveness of each component as well as the self-review mechanism, while more detailed analysis further illustrate the evolution of experiential memory throughout practice process and reveal the important roles of the elaborate designs in the PraMem. 
The main contributions of this paper can be summarized as:
\begin{itemize}[leftmargin=1em,itemindent=0em,itemsep=0pt,topsep=0pt]
\item We propose PraMem, which conducts iterative practice over the lengthy historical sequence to build an experiential memory, thereby assisting for accurate long-horizon behavior prediction.
\item We design the self-review mechanism, which can filter out reflective proposals without sufficient groundedness and good generalizability, thereby ensuring the overall effectiveness of PraMem.
\item We conduct extensive experiments and detailed analysis, which confirm the advantage of PraMem and provide several valuable insights into the evolution and mechanism of the experiential memory.
\end{itemize}
\section{PraMem}

To tackle the aforementioned challenges of LLM-based long-horizon behavior prediction, as illustrated in Figure~\ref{fig:method}, we draw inspiration from dialectical philosophy and propose PraMem, which conducts beforehand practice over the lengthy historical sequence to build a user-specific and time-evolving experiential memory, thereby serving as additional input for accurate prediction.

\subsection{Task Formulation}

\textbf{Vanilla Process.} From the perspective of a single user, the LLM is employed to predict the user's action $\hat{a}_t$ in a current scene $s_t$ based on the user's lengthy historical sequence of scene-action records:
\begin{equation}
    \hat{a}_t = \mathrm{LLM}\!\left( \mathcal{H}_t,\, s_t \right), \quad \mathcal{H}_t = \{(s_i, a_i)\}_{i=1}^{t-1}
\end{equation}
where $\mathcal{H}_t$ denotes the historical sequence up to step $t$, with $s_i$ and $a_i$ representing the scene and action at step $i$. All scenes and actions are represented in natural language, and the LLM generates $\hat{a}_t$ in an autoregressive manner conditioned on the textual concatenation of $\mathcal{H}_t$ and $s_t$, with the goal of unbiasedly capturing the user's underlying behavioral patterns from the lengthy historical sequence.

\textbf{PraMem Process.} To boost LLMs on long-horizon behavior prediction, PraMem performs beforehand practice over the lengthy historical sequence to build a user-specific and time-evolving experiential memory $\mathcal{M}_t$, which can assist the LLM in accurate prediction, as following formula:
\begin{equation}
\label{for:2}
    \mathcal{M}_t = \mathrm{Practice}\!\left(\mathcal{H}_t,\, \mathcal{M}_{t'}\right), \quad \hat{a}_t = \mathrm{LLM}\!\left( \tilde{\mathcal{H}}_t,\, s_t,\, \mathcal{M}_t \right), \quad \tilde{\mathcal{H}}_t = \{(s_i, a_i)\}_{i=t-k}^{t-1}
\end{equation}
where $\mathcal{M}_{t'}$ ($t' < t$) denotes the previous state of the experiential memory, based on which $\mathcal{M}_t$ is incrementally evolved by practicing on the newly accumulated records, and $\tilde{\mathcal{H}}_t$ is a short suffix containing only the $k$ most recent records adjacent to the current prediction step ($k \ll t$).

\subsection{Practice-derived Experiential Memory}

To comprehensively address the two core challenges of LLM-based long-horizon behavior prediction, PraMem dynamically maintains an experiential memory that consists of two types of experience:

\begin{itemize}[leftmargin=1em,itemindent=0em,itemsep=0pt,topsep=0pt]
\item \textbf{Pattern Experience (Pat. E.)}: Presenting user behavioral patterns embedded in the lengthy historical sequence, for example, \textit{this user likes Apple's electronic products, this user buys daily necessities at the beginning of each month, this user tends to add items to favorites before purchase}.
\item \textbf{Bias-alert Experience (Bia. E.)}: Alerting the LLM to intrinsic biases that are likely to arise when handling the current user, for example, \textit{avoiding the proximal bias that only considers recent behavioral records, avoiding the prejudice assumption that Asians prefer some mysticism things}.
\end{itemize}

With this twofold experiential memory, the LLM can acquire reliable behavioral patterns of the user and stay alert to its own error-prone biases before making the formal prediction, thereby enabling more accurate long-horizon behavior prediction. 
Notably, the experiential memory is user-specific and time-evolving: as the user's behavioral sequence continues to grow, PraMem dynamically maintains the experiential memory by iteratively performing existing experience trial, reflective proposal generation, and consensual experience adjustment, so that the memory becomes increasingly precise in capturing behavioral patterns and increasingly pertinent in signaling cognitive biases.

\textbf{Existing Experience Trial.} 
In order to sufficiently expose shortcomings of the existing experience, PraMem samples $q$ segments from the historical sequence to construct behavior prediction practice samples with ground-truth labels, and then prompts the LLM to carefully do explicit deep thinking under the current experiential memory so as to fully leverage the existing experience for prediction.
Specifically, let $\mathcal{M}_{t'}$ denote the current state of the experiential memory, which has been built upon the historical sequence $\mathcal{H}_{t'}$ up to step $t'$. Taking a single sample as an illustration:
\begin{equation}
    (\tilde{s}_j, \tilde{a}_j) \sim \mathcal{H}_t \setminus \mathcal{H}_{t'}, \quad \tilde{\mathcal{H}}_j = \{(s_i, a_i)\}_{i=j-k}^{j-1}, \quad (\hat{a}_j, r_j) = \mathrm{LLM}_{\text{thinking}}\!\left(\tilde{\mathcal{H}}_j,\, \tilde{s}_j,\, \mathcal{M}_{t'}\right)
\end{equation}
where $(\tilde{s}_j, \tilde{a}_j)$ denotes the sampled target scene and its ground-truth label, with $j$ later than the time point $t'$ covered by the current experiential memory (i.e., $t' < j < t$), ensuring that the evolution of the experiential memory remains consistent with the temporal order of the user's behavioral sequence,
$\tilde{\mathcal{H}}_j$ is the short historical suffix preceding the target scene, which together with $\tilde{s}_j$ constitutes the complete practice sample, $k$ is a hyperparameter controlling the length of $\tilde{\mathcal{H}}_j$, $\hat{a}_j$ and $r_j$ are respectively the predicted action and the explicit deep-thinking process generated by the LLM under the guidance of the current experiential memory. Detailed prompts used in this part are shown in Appendix~\ref{app:Prompt1}.

\textbf{Reflective Proposal Generation.}
After the above trial on the sampled practice sample, PraMem performs reflection by jointly considering the practice sample, deep-thinking process, and comparison between predicted action and ground-truth label, thereby producing proposals for revising, supplementing, or pruning the experiential memory. 
Specifically, each proposal is composed of the type of experience it targets and detailed content describing the specific operation. 
To ensure the reliability of the produced proposals, a self-review mechanism is further introduced to filter out the unreliable ones, and only those passing the review are accepted into a proposal pool that supports the subsequent adjustment of the experiential memory.
The above process can be expressed as following:
\begin{equation}
\label{for:4}
    \{p_i\}_{i=1}^{n} = \mathrm{LLM}_{\text{reflection}}\!\left(\tilde{\mathcal{H}}_j,\, \tilde{s}_j,\, \tilde{a}_j,\, \hat{a}_j,\, r_j\right), \quad \mathcal{P} \leftarrow \mathcal{P} \cup \{p_i \mid \mathrm{SelfReview}(p_i) = \text{Pass}\}_{i=1}^{n}
\end{equation}
where $p_i$ denotes the $i$-th proposal generated by the LLM after careful reflection, and $n$ is a hyperparameter specifying the number of generated proposals. $\mathrm{SelfReview}(\cdot)$ judges whether a proposal is well-grounded and generalizable, and $\mathcal{P}$ is the proposal pool that accumulates all reliable proposals across rounds for the subsequent experience adjustment. The detailed design of self-review mechanism is presented in Section~\ref{sec:self} and detailed prompts used in this part are shown in Appendix~\ref{app:Prompt2}.

\textbf{Consensual Experience Adjustment.}
In the iterative loop, the above existing experience trial and reflective proposal generation are executed cyclically to accumulate the proposal pool, while the experience adjustment is triggered every $T$ rounds to avoid instability caused by occasional user behaviors in the historical sequence. 
Specifically, when triggered, PraMem utilize the LLM to perform consensus-driven adjustment of experiential memory based on the proposal pool, where only operations consistently supported by multiple proposals are adopted, enabling stable evolution of experiential memory. The consensus-driven adjustment process can be formulated as following:
\begin{equation}
    \mathcal{M}_t,\, \mathcal{P} = \mathrm{LLM}_{\text{consensus}}\!\left(\mathcal{M}_{t'},\, \mathcal{P}\right)
\end{equation}
where $\mathcal{M}_t$ is the updated experiential memory produced by the LLM through identifying and applying the consistently supported operations in $\mathcal{P}$. After the adjustment, the proposals that have been adopted are removed from $\mathcal{P}$, while the remaining ones are retained for accumulation in the subsequent iterative loop. Detailed prompts used in this part are shown in Appendix~\ref{app:Prompt3}.

\section{Self-review Mechanism}
\label{sec:self}
\begin{figure*}[t]
\centering
 \includegraphics[width=\linewidth]{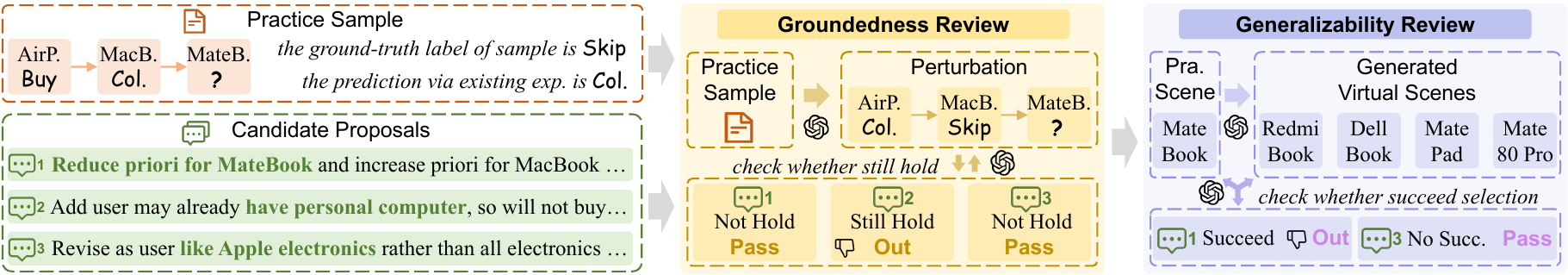} 
\caption{Self-review mechanism only retains reliable proposals that are both genuinely grounded in and generalizable beyond the practice sample, thereby ensuring the overall effectiveness of PraMem.}
\label{fig:self}
\end{figure*}
In the above PraMem, reflective proposals directly dictate the generation and evolution of the experiential memory, therefore, the quality of reflective proposals serves as a critical determinant of overall effectiveness. 
On one hand, proposals must be well-grounded in the practice sample rather than being fabricated by the LLM, thereby ensuring the resulting experiential memory faithfully captures authentic requirements for prediction.
On the other hand, proposals must be sufficiently generalizable rather than being overly specific to a single practice sample, thereby enabling the experiential memory to effectively guide predictions on unseen scene. 
To this end, we introduce a self-review mechanism to sequentially conduct groundedness review and generalizability review, which can effectively filter out proposals that lack substantial groundedness or possess poor generalizability, as in Figure~\ref{fig:self}.

\textbf{Groundedness Review.}
To identify proposals that lack groundedness with respect to the practice sample, we leverage the LLM to perturb the historical sequence of the sample such that its underlying behavioral patterns are substantially altered. 
Then the LLM is employed to assess whether the original proposal still holds on the perturbed practice sample. If still holding, it indicates that the proposal is not genuinely grounded in the practice sample. This process can be formulated as follows:
\begin{equation}
    \tilde{\mathcal{H}}_j^{\,\prime} = \mathrm{LLM}_{\text{perturb}}(\tilde{\mathcal{H}}_j), \quad
    \mathrm{Gr.Review}(p_i) = 
    \begin{cases}
        \text{Out}, & \text{if}\; \mathrm{LLM}_{\text{judge}}(p_i~|~ \tilde{\mathcal{H}}_j^{\,\prime},\, \tilde{s}_j,\, \tilde{a}_j) = \text{Hold} \\
        \text{Pass}, & \text{otherwise}
    \end{cases}
\end{equation}
where $\tilde{\mathcal{H}}_j^{\,\prime}$ is the  perturbed historical sequence, in which the actions are deliberately rewritten by the LLM to substantially change the behavioral pattern. Prompts for this part are in Appendix~\ref{app:Prompt4}.

\textbf{Generalizability Review.}
To identify proposals that lack generalizability, we leverage the LLM to rewrite the prediction scene of sample, generating virtual scenes that share the same  category as the true scene but differs in fine-grained attributes, thereby constituting a multiple-choice task. 
Then the LLM is employed to determine whether the proposal can unambiguously guide selection of true scene. If so, the proposal is overly specific to the current practice sample. The formula is as follows:
\begin{equation}
\label{for:7}
\{s_v^{(i)}\}_{i=1}^{m} = \mathrm{LLM}_{\text{rewrite}}(\tilde{s}_j), \quad
\mathrm{Ge.Review}(p_i) = 
\begin{cases}
    \text{Out}, & \text{if}\; \mathrm{LLM}_{\text{select}}(\mathcal{S}_j~|~p_i) = \tilde{s}_j \\
    \text{Pass}, & \text{otherwise}
\end{cases}
\end{equation}
where $\{s_v^{(i)}\}_{i=1}^{m}$ denotes the $m$ virtual scenes based on the true scene $\tilde{s}_j$, and $\mathcal{S}_j = \{\tilde{s}_j\} \cup \{s_v^{(i)}\}_{i=1}^{m}$ is the scene set forming the multiple-choice task. Prompts for this part are in Appendix~\ref{app:Prompt5}.

\section{Experiments}

\subsection{Experimental Settings}

\textbf{Datasets.}
We select two publicly available and widely recognized benchmarks that encompass diverse long-horizon behavior prediction tasks across multiple scenarios, enabling a comprehensive and rigorous evaluation.
(1) \textbf{OminiBehavior}~\citep{chen2026towards}, which contains long historical sequences of users from the Kuaishou platform, requiring the behavior prediction across four scenarios: video browsing, live streaming, advertisement, and e-commerce.
We adopt the behavior prediction tasks from the benchmark and follow its standard evaluation metrics, namely accuracy (ACC) and F1-score, with test instances formulated as binary classification tasks such as whether a user will favorite a video, follow the video author, send gifts in live streams, comment in live streams, click ads, and make purchases.
(2) \textbf{MovieLens-1M}~\citep{harper2015movielens}, which contains extensive user rating records in chronological order for movies on a 5-point scale (1 to 5 stars), formulating a 5-class classification task. 
We filter users with more than 300 rating records and construct long-horizon behavior prediction tasks, where the goal is to predict a user's rating for the current movie based on the lengthy historical sequence. We present results using confusion matrix and compare the performance of different methods based on the diagonal dominance of different matrices.

\textbf{Baselines.}
We select some commonly used methods as well as advanced memory management approaches for long-horizon behavior prediction tasks to conduct comprehensive comparisons with PraMem.
(1) \textbf{No Memory Management}. This category includes methods that directly set the LLM's input window to 32K tokens (Long-context) to accommodate as much of the long historical sequence as possible, as well as Truncation methods that retain only the most recent records from the long historical sequence.
(2) \textbf{Simple Memory Management}. This category includes the RAG method~\citep{lewis2020retrieval} that uses a retriever to recall the most relevant records as memory from the long historical sequence with respect to the prediction scene, and summary method~\citep{wang2023recursively} that constructs memory through iterative summarization based on the long historical sequence.
(3) \textbf{Advanced Memory Management}. This category includes Mem0~\citep{chhikara2025mem0} and MemOS~\citep{li2025memos}, which leverage LLMs to perform information extraction and summarization on the long historical sequence and use a retriever to recall the most relevant text as memory, as well as ProEx~\citep{zhang2026proex}, which employs LLMs to induce user behavioral patterns from the long historical sequence to generate multiple dynamic  profiles for each user, and refines them by cross-validation across profiles to obtain the most accurate user profile as memory.

\textbf{Implementation Details.}
We store the experiential memory in local files, with each user corresponding to a JSON file that contains three lists, representing the pattern experience, bias-alert experience, and proposal pool, respectively, and each experience or proposal is an element of the list.
During initialization, the pattern experience is initialized with basic user information explicitly provided in the historical sequence of the benchmark, the bias-alert experience is initialized with generic prompts regarding common LLM biases, and the proposal pool is left empty.
During the experiential memory maintenance process, we employ GPT-OSS-120B\footnote{https://huggingface.co/openai/gpt-oss-120b} as the LLM for all procedures in PraMem. The hyperparameter $n$ in Formula~\ref{for:4} is set as 10, the $m$ in Formula~\ref{for:7} is set as 7, the interval rounds $T$ to trigger consensual experience adjustment is set as 5, the practice sample number $q$ in each round is set as 1, and totally PraMem iteratively performs 80 rounds to dynamically maintain the experiential memory for each user.  
For evaluation and comparison between baselines and PraMem, the input window of Long-context is set to 32K tokens, while that of Truncation is set to 8K tokens to retain the most recent $k$ records. All other baselines and PraMem do same as Truncation, taking only these $k$ records as input, where $k$ corresponds to that in Formula~\ref{for:2}. To ensure a fair comparison, the memory is uniformly concatenated with the truncated historical sequence as the joint input for PraMem and all memory-based baselines. And we employ two backbones including GPT-OSS-120B\footnotemark[1] and Qwen3.5-35B-A3B\footnote{https://huggingface.co/Qwen/Qwen3.5-35B-A3B} in evaluation for credibility.
In experiments, we deploy all LLMs as API via vLLM\footnote{https://pypi.org/project/vllm/} for convenient access, and keeping all default parameters to ensure reproducibility.

\subsection{Main Results}

\begin{figure*}[t]
\vspace{-10pt}  
\centering
 \includegraphics[width=\linewidth]{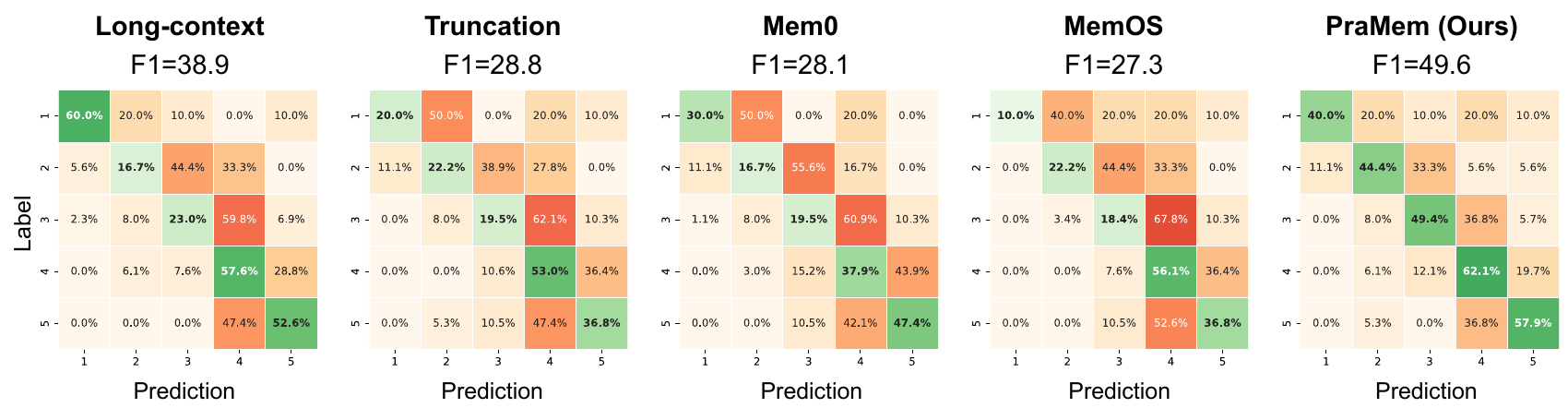} 
\caption{Main results by the confusion matrix on MovieLens-1M dataset. PraMem displays stronger diagonal dominance than baselines, confirming its advantage on long-horizon behavior prediction.}
\label{fig:ml}
\vspace{-10pt}  
\end{figure*}

Results compared with baselines are shown in Table~\ref{table:main_results} and Figure~\ref{fig:ml}, there are two main conclusions:

\textbf{PraMem is a powerful method for addressing long-horizon behavior prediction tasks}. As shown in Table~\ref{table:main_results}, on OmniBehavior, PraMem surpasses the baselines on the vast majority of tasks and metrics, achieving the best overall performance, improving the ACC from 73.5 (Truncation) to 84.7 and the F1 from 24.7 to 31.6. In contrast, prior advanced memory management methods yield marginal gains: Mem0, which relies on LLM-based complex extraction, aggregation, and retrieval, achieves an ACC of 74.7 and an F1 of 26.4, MemOS achieves 73.1 and 26.2 in ACC and F1, respectively, and ProEx, which performs memory management via dynamic user profiles generated by LLMs, achieves 77.7 and 26.9 in ACC and F1. In addition, as illustrated in Figure~\ref{fig:ml}, on MovieLens-1M, the confusion matrix of PraMem exhibits stronger diagonal concentration than baselines, indicating more accurate predictions and further confirming the effectiveness of PraMem.

\textbf{The practice-derived experiential memory generalizes across different LLMs, demonstrating good practicality.} As described in the implementation details, we employ only GPT-OSS-120B for constructing the experiential memory, while adopting two backbones during evaluation. As shown in Table~\ref{table:main_results}, the experiential memory is effective not only on GPT-OSS-120B but also yields substantial performance gains when the backbone is switched to Qwen3.5-35B-A3B. This observation aligns with design principles and prior findings: the pattern experience primarily presents user behavioral patterns, which are naturally transferable across backbones, as for the bias-alert experience, prior studies have revealed that different LLMs share certain common internal cognition~\citep{huh2024platonic}, so that bias-alert experience derived from one LLM remains effective for another. This  generalizability offers good practical value, as the experiential memory can be constructed once and directly reused across different backbones, eliminating the need for model-specific reconstruction.

\begin{table*}[t]
\centering
\renewcommand{\arraystretch}{1.05}
\resizebox{\textwidth}{!}{
\begin{tabular}{ll *{10}{c}}
\toprule
\multirow{2}{*}{\textbf{Category}} & \multirow{2}{*}{\textbf{Method}} &
\multicolumn{2}{c}{\textbf{Video}} &
\multicolumn{2}{c}{\textbf{Live}} &
\multicolumn{2}{c}{\textbf{Ads}} &
\multicolumn{2}{c}{\textbf{E-commerce}} &
\multicolumn{2}{c}{\textbf{Overall}} \\
\cmidrule(lr){3-4}\cmidrule(lr){5-6}\cmidrule(lr){7-8}\cmidrule(lr){9-10}\cmidrule(lr){11-12}
& & ACC & F1 & ACC & F1 & ACC & F1 & ACC & F1 & ACC & F1 \\
\midrule
\rowcolor[rgb]{.90,.90,.90}
\multicolumn{12}{l}{\textit{Backbone: GPT-OSS-120B}} \\
\midrule
\multirow{2}{*}{No Memory}
 & Long-context & 74.9 & 23.4 & 67.2 & 25.2 & 59.4 & 28.3 & 72.0 & 21.4 & 68.4 & 24.6 \\
 & Truncation   & 79.5 & 23.9 & 73.2 & 27.4 & 61.0 & 26.2 & 80.1 & 21.3 & 73.5 & 24.7 \\
\cmidrule(lr){1-12}
\multirow{2}{*}{Simple Memory}
 & RAG~\citep{lewis2020retrieval}     & 77.9 & 20.6 & 71.6 & 26.6 & 61.3 & 22.2 & 65.7 & 14.3 & 69.1 & 20.9 \\
 & Summary~\citep{wang2023recursively} & 80.1 & 23.2 & 77.0 & 28.4 & 64.7 & 27.5 & 74.5 & 18.2 & 74.1 & 24.3 \\
\cmidrule(lr){1-12}
\multirow{3}{*}{Advanced Memory}
 & Mem0~\citep{chhikara2025mem0}    & 81.7 & 25.9 & 75.0 & 28.5 & 63.1 & 29.5 & 78.9 & 21.6 & 74.7 & 26.4 \\
 & MemOS~\citep{li2025memos}   & 78.9 & 25.0 & 74.4 & 28.8 & 61.8 & 29.0 & 77.3 & 22.0 & 73.1 & 26.2 \\
 & ProEx~\citep{zhang2026proex} & 82.5 & \textbf{26.2} & 77.1 & 27.7 & 68.1 & 28.5 & 83.3 & 25.2 & 77.7 & 26.9 \\
\cmidrule(lr){1-12}
Experiential Memory
 & \textbf{PraMem (Ours)} & \textbf{86.4}* & 25.6* & \textbf{83.7}* & \textbf{29.9}* & \textbf{82.3}* & \textbf{43.7}* & \textbf{86.2}* & \textbf{27.1}* & \textbf{84.7}* & \textbf{31.6}* \\
\midrule
\rowcolor[rgb]{.90,.90,.90}
\multicolumn{12}{l}{\textit{Backbone: Qwen3.5-35B-A3B}} \\
\midrule
\multirow{2}{*}{No Memory}
 & Long-context & 61.5 & 18.1 & 62.0 & 22.5 & 55.7 & 25.4 & 62.2 & 20.1 & 60.3 & 21.5 \\
 & Truncation   & 67.7 & 18.6 & 64.3 & 22.6 & 58.0 & 24.4 & 65.3 & 18.1 & 63.8 & 20.9 \\
\cmidrule(lr){1-12}
\multirow{2}{*}{Simple Memory}
 & RAG~\citep{lewis2020retrieval}     & 64.2 & 17.0 & 60.5 & 19.7 & 55.9 & 25.1 & 61.3 & 15.8 & 60.5 & 19.4   \\
 & Summary~\citep{wang2023recursively} & 67.3 & 19.2 & 65.6 & 22.9 & 59.4 & 25.7 & 64.2 & 16.5 & 64.1 & 21.1 \\
\cmidrule(lr){1-12}
\multirow{3}{*}{Advanced Memory}
 & Mem0~\citep{chhikara2025mem0}    & 67.2 & 19.0 & 66.6 & 24.4 & 55.6 & 24.0 & 65.9 & 19.3 & 63.8 & 21.7 \\
 & MemOS~\citep{li2025memos}   & 65.8 & 19.5 & 65.6 & 24.2 & 54.2 & 24.6 & 67.9 & 19.4 & 63.4 & 21.9 \\
 & ProEx~\citep{zhang2026proex} & 70.7 & 18.7 & 69.0 & 23.4 & 61.3 & 24.0 & 69.2 & 19.8 & 67.6 & 21.5 \\
\cmidrule(lr){1-12}
Experiential Memory
 & \textbf{PraMem (Ours)} & \textbf{77.7}* & \textbf{20.3}* & \textbf{75.6}* & \textbf{26.9}* & \textbf{74.6}* & \textbf{32.4}* & \textbf{74.3}* & \textbf{21.5}* & \textbf{75.6}* & \textbf{25.3}* \\
\bottomrule
\end{tabular}}
\caption{Main results across two backbones and four scenarios from OminiBehavior benchmark. PraMem is effective for long-horizon behavior prediction, and the experiential memory can transfers well across  backbones. * indicates statistically significant improvements (p < 0.01) over baselines.}
\label{table:main_results}
\end{table*}

\begin{figure*}[t]
\centering
 \includegraphics[width=\linewidth]{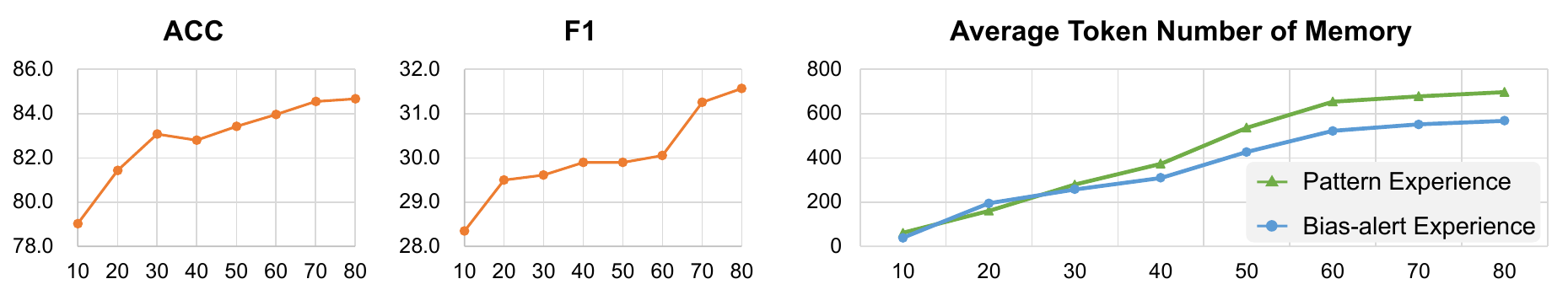} 
\caption{Evolution of experiential memory. The overall performance improves with more practice rounds,  while the average token number of experiential memory first increases and then stabilizes.}
\label{fig:evloution}
\vspace{-10pt}  
\end{figure*}
\subsection{Evolution of Experiential Memory}

To gain deeper insights into the role of experiential memory and its evolution throughout the iterative practice process, we present in Figure~\ref{fig:evloution} the performance achieved using experiential memory from different practice rounds, along with statistics on the average token number of experiential memory. There are two conclusions about evolution of experiential memory during practice process:

\textbf{The performance  of experiential memory on long-horizon behavior prediction progressively improves with more practice rounds.} As shown in the leftmost subplot of Figure~\ref{fig:evloution}, using the experiential memory obtained after 10 practice rounds for long-horizon behavior prediction can achieve an overall score of 79.0 ACC and 28.3 F1, which further improves to 83.5 ACC and 30.7 F1 at the 40th round, and reaches 84.7 ACC and 31.6 F1 at the 80th round. This monotonically increasing overall score with respect to the number of practice rounds suggests that the experiential memory undergoes a progressive evolution throughout the iterative practice process, continuously enhancing its utility for long-horizon behavior prediction. This phenomenon is consistent with our design, where iterative trial-and-error enables the gradual derivation of a good experiential memory.

\textbf{During evolution, the total length of experiential memory first increases and then stabilizes.} As shown in the rightmost subplot of Figure~\ref{fig:evloution}, the total length of experiential memory gradually increases from round 10 to round 60, and then stabilizes after round 60, while the overall performance  exhibits a steady upward trend throughout the entire process. This is because PraMem adjusts the experiential memory through three types of operations (i.e. revising, supplementing, and pruning). After a certain number of rounds, the memory achieves relatively comprehensive coverage, and subsequent iterations focus on refining and correcting the existing experience rather than continuously adding new information. This further demonstrates the practical value of PraMem: unlike conventional memory management methods that always keep accumulating information, PraMem avoids length explosion as rounds increase, thereby preventing additional long-context burdens on the LLM.



\begin{wraptable}{r}{0.50\textwidth}
\centering
\vspace{-8pt}
\renewcommand{\arraystretch}{1.05}
\setlength{\tabcolsep}{4pt}
\resizebox{0.50\textwidth}{!}{
\begin{tabular}{lccccc}
\toprule
\textbf{Method} & \textbf{Video} & \textbf{Live} & \textbf{Ads} & \textbf{E-commerce} & \textbf{Overall} \\
\midrule
\textbf{PraMem (Ours)} 
& 25.6 & \textbf{29.9} & \textbf{43.7} & \textbf{27.1} & \textbf{31.6} \\
~~~~w/o deep-thinking trial   
& 25.5 & 26.2 & 41.2 & 24.1 & 29.3 \\
~~~~w/o reflective proposal   
& 23.9 & 29.5 & 43.5 & 27.0 & 30.9 \\
~~~~w/o consensual adjustment   
& \textbf{26.0} & 28.2 & 38.3 & 22.6 & 28.8 \\
\bottomrule
\end{tabular}}
\caption{Ablation results in terms of  F1. Deep thinking in experience trial, reflection in proposal generation, and consensus setting in experience adjustment all play crucial positive effect.}
\label{table:ab_results}
\vspace{-10pt}
\end{wraptable}

\subsection{Detailed Analysis}

\textbf{Ablation Results.}
To investigate the necessity of several key designs, we conduct ablation studies  in Table~\ref{table:ab_results}. Specifically, ``w/o deep-thinking trial'' is removing  deep thinking in experience trial and instead making predictions directly, which fails to adequately expose shortcomings. ``w/o reflective proposal'' is removing reflection in proposal generation and instead generating proposals directly from prediction-label comparison of the practice sample, which tends to produce inaccurate proposals. ``w/o consensual adjustment'' is removing consensus setting in experience adjustment and instead directly adopting all proposals, which degrades quality of  experiential memory due to occasional behaviors. The results show \textit{deep thinking in experience trial, reflection in proposal generation, and consensus setting in experience adjustment all play crucial positive roles.}  

\begin{figure}[H]
\vspace{-10pt}  
    \centering
    \begin{minipage}{0.48\linewidth}
        \centering
        \includegraphics[width=\linewidth]{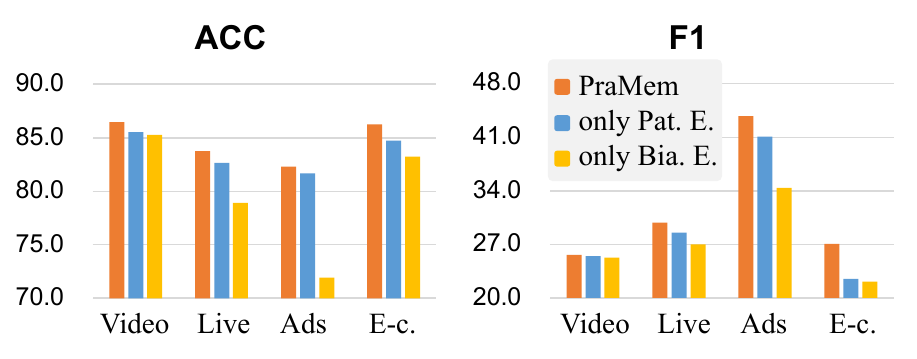}
        \caption{Both of the pattern experience and the bias-alert experience play positive roles.}
        \label{fig:twofold}
    \end{minipage}
    \hfill
    \begin{minipage}{0.48\linewidth}
        \centering
        \includegraphics[width=\linewidth]{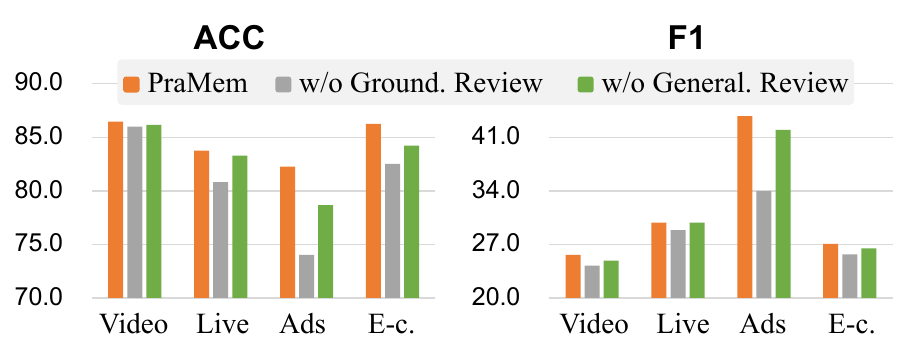}
        \caption{The self-review mechanism effectively ensures  quality of proposals and experiences.}
        \label{fig:selfreview}
    \end{minipage}
\vspace{-10pt}  
\end{figure}

\textbf{Effectiveness of Twofold Experience.}
To in-detail validate the contributions of the two types of experience in PraMem (i.e., pattern experience and bias-alert experience), we report in Figure~\ref{fig:twofold} the performance achieved using only one type of experience. The results show that removing either type leads to a performance decline, which to some extent confirms that prior memory management works, which solely consider the historical sequence while neglecting the intrinsic biases of the LLM, indeed suffer from a notable limitation. In contrast, \textit{PraMem jointly considers pattern experience for presenting user behavioral patterns and bias-alert experience for alerting against LLM intrinsic biases, representing a more advanced memory strategy for long-horizon behavior prediction}.

\textbf{Effectiveness of Self-review Mechanism.}
To in-detail validate contributions of the self-review mechanism in PraMem (i.e., groundedness review and generalizability review), we report in Figure~\ref{fig:selfreview} the performance when removing the two kind of self-review components. The results show that both groundedness review and generalizability review contribute positively to the overall performance for long-horizon behavior prediction, demonstrating that \textit{the self-review mechanism can effectively identify and filter out unreliable proposals, thereby ensuring the quality of the experiential memory}.

\textbf{Case Study.}
To intuitively demonstrate the advantages of the experiential memory built by PraMem over traditional memory management methods, we conduct a case study that compares the memory maintained for the same user under PraMem and the representative baseline MemOS, as shown in Appendix~\ref{app:case}.
This case reveals a fundamental difference between the two: \textit{the baseline's memory is a record of facts, remaining at the level of what has happened, in contrast, PraMem's memory is the explicit and concise experience distilled through iterative practice}, which can effectively presents the user's latent behavioral patterns and pre-alerts the LLM against its systematic biases, thereby genuinely resolving the core challenges of long-horizon behavior prediction for better performance.

\section{Related Work}

\textbf{Experiential Memory.}
Experiential memory, as a core mechanism for self-evolution in agents, has been widely adopted in the agent domain~\citep{hu2025memory,gao2025survey}. The core idea is to encode historical interaction trajectories, successful experiences, and failure lessons into durable, retrievable knowledge representations, enabling agents to reuse such experiences in subsequent tasks to avoid repeating mistakes and to improve decision quality~\citep{zheng2023synapse,zhang2025memevolve}.
Specifically, some methods consolidate experience into executable procedural capacities such as building ever-growing skill libraries~\citep{wang2023voyager,zheng2025skillweaver,bouzenia2025repairagent,fang2025memp}.
Other methods distill transferable reasoning patterns and high-level insights from past trajectories to guide subsequent decisions~\citep{ouyang2025reasoningbank,li2026digimongpt,zhu2026re,yang2026plugmem}.
Long-horizon behavior prediction tasks inherently contain abundant practice samples, providing an ideal data foundation for constructing experiential memory. Therefore, introducing experiential memory mechanisms into this task holds great promise and potential.

\textbf{Memory for Behavior Prediction.}
To assist LLMs in long-horizon behavior prediction, prior works have primarily focused on memory management through extraction, summarization, and retrieval over the historical sequence, thereby alleviating the long-context burden of LLMs~\citep{zheng2024harnessing,wang2024recmind,xi2024memocrs,wang2025user,huang2025mr}, but do not directly address the two core challenges of long-horizon behavior prediction. 
More recently, some works have attempted to dynamically construct user profiles as a form of memory, such as ProEx and LettinGo build comprehensive user profiles from multiple aspects by jointly considering the historical sequence through multi-granularity consistency checks~\citep{wang2025lettingo,zhang2026proex}, and 
DGDPO performs diagnostic evaluation on the user profile by sampling segments from the historical sequence, so as to identify and correct deficiencies in the current profile~\citep{liu2026diagnostic}. 
Nevertheless, these works focus solely on memory management over the historical sequence and overlook the intrinsic deficiencies exhibited by LLMs when performing the task, and they lack a review mechanism for profile operations, making it difficult to guarantee the reliability of the memory maintenance process.

\section{Discussion}

\textbf{Conclusion.} In this paper, we draw inspiration from dialectical philosophy and propose PraMem, which conducts beforehand practice over the lengthy historical sequence to build an experiential memory, thereby assisting for accurate long-horizon behavior prediction. The PraMem iteratively performs existing experience trial, reflective proposal generation, and consensual experience adjustment to maintain the pattern experience for presenting user behavioral patterns and bias-alert experience for alerting LLM intrinsic biases. And we design the self-review mechanism, which can filter out reflective proposals without sufficient groundedness and good generalizability, thereby ensuring the overall effectiveness of PraMem. Experiments on various long-horizon behavior prediction tasks demonstrate PraMem is a powerful method with better overall performance than all baselines, and provide some valuable insights of experiential memory. Therefore, this paper offers a promising direction for more powerful memory systems for long-horizon behavior prediction in the future.

\textbf{Limitations.}
A main limitation of this paper lies in the memory construction cost of PraMem. Compared with conventional memory management methods based on information extraction, aggregation, and retrieval, the iterative practice process of PraMem inevitably requires more time to build the experiential memory for better performance. Although this process is conducted offline and thus does not affect the efficiency of the actual prediction, it remains a non-trivial limitation that calls for future investigation, for instance through more efficient practice scheduling or parallelized trial-and-reflection strategies.
Potential negative societal impact is in Appendix~\ref{app:social}.

\newpage

\bibliography{neurips_2026}
\bibliographystyle{neurips_2026}

\newpage

\appendix

\section{Prompt for Existing Experience Trial}
\label{app:Prompt1}
\begin{figure*}[t]
\centering
 \includegraphics[width=\linewidth]{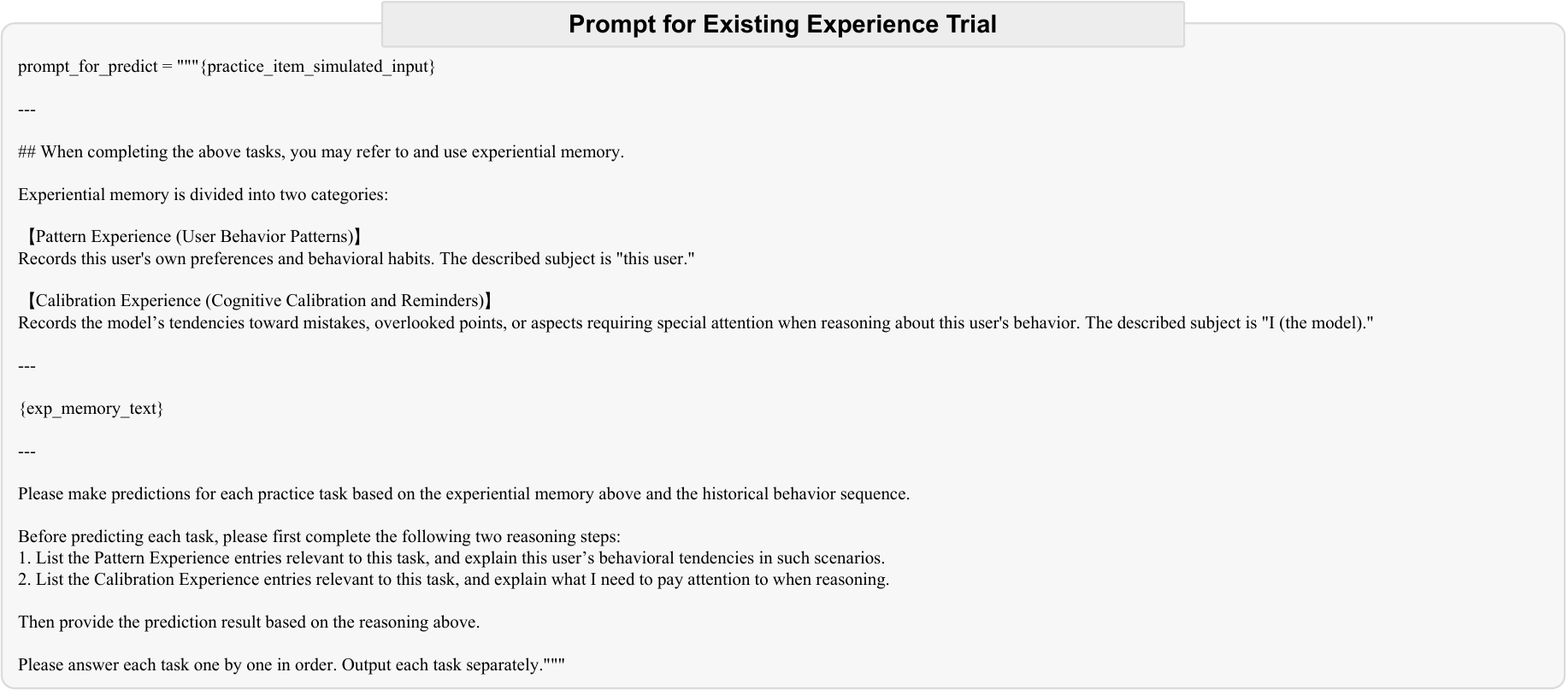} 
\caption{Prompt for existing experience trial.}
\label{fig:prompt1}
\end{figure*}

This prompt is used in the existing experience trial stage of PraMem, where the LLM performs beforehand practice on constructed practice samples from the historical sequence. Given the current experiential memory, the LLM is required to predict the action for each practice task based on both the historical behavior sequence and the accumulated experience. To make the use of experiential memory explicit and controllable, the prompt asks the LLM to separately identify relevant pattern experience, which captures this user’s behavioral patterns, and relevant deficiency experience, which alerts the model to its own possible intrinsic deficiencies when reasoning about this user. This explicit deep-thinking process helps expose whether the current memory can effectively support prediction and provides necessary reasoning traces for subsequent reflection. The prompt is in Figure~\ref{fig:prompt1}.

\section{Prompt for Reflective Proposal Generation}
\label{app:Prompt2}
\begin{figure*}[t]
\centering
 \includegraphics[width=\linewidth]{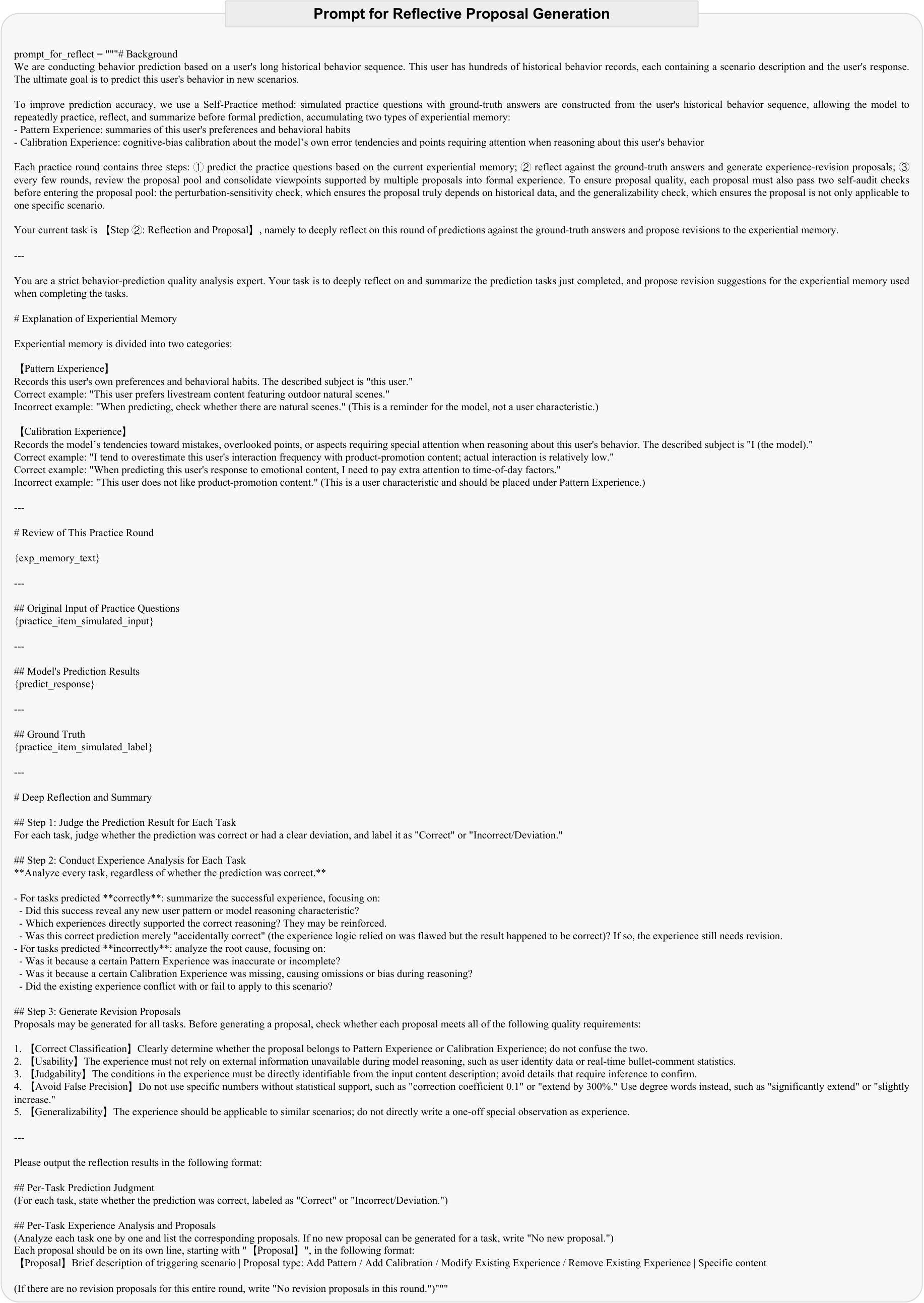} 
\caption{Prompt for reflective proposal generation.}
\label{fig:prompt2}
\end{figure*}
This prompt is used in the reflective proposal generation stage of PraMem, where the LLM reflects on the just-completed practice round by comparing its predictions with the ground-truth labels. Its purpose is to transform trial-and-error outcomes into candidate revisions for experiential memory, so that the model can gradually induce reliable behavioral patterns and identify its own deficiencies for the current user. The prompt requires the LLM to analyze both correct and incorrect predictions, since correct predictions may reinforce useful experience or expose accidental correctness, while incorrect predictions can reveal missing, inaccurate, or conflicting experience. It also explicitly distinguishes pattern experience, which describes this user’s behavioral patterns, from deficiency experience, which records the model’s possible mistakes or overlooked aspects, ensuring that reflective proposals are assigned to the proper memory category. The prompt is shown in Figure~\ref{fig:prompt2}.

\section{Prompt for Consensual Experience Adjustment}
\label{app:Prompt3}
\begin{figure*}[t]
\centering
 \includegraphics[width=\linewidth]{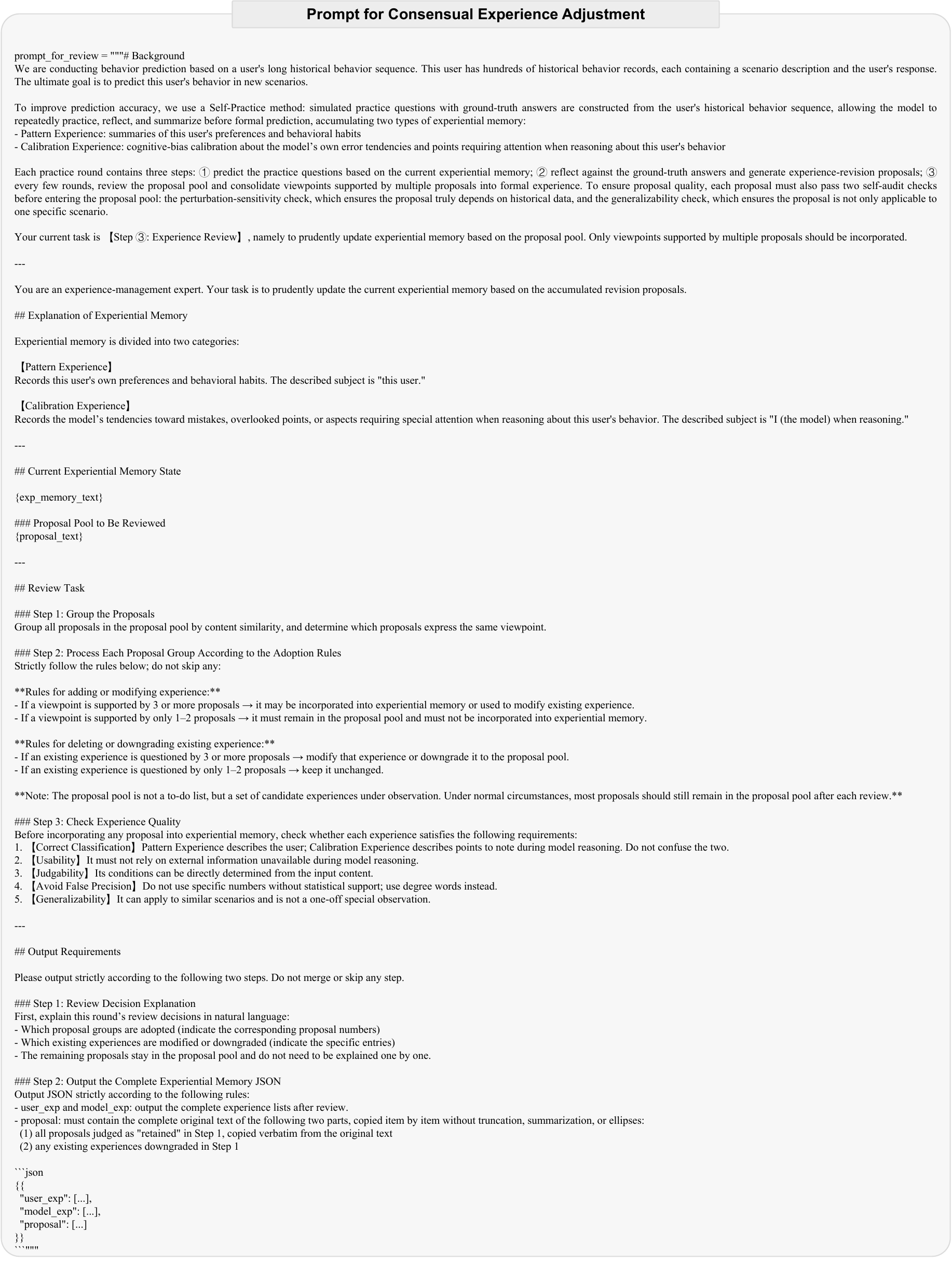} 
\caption{Prompt for consensual experience adjustment.}
\label{fig:prompt3}
\end{figure*}
This prompt is used in the consensual experience adjustment stage of PraMem, where the accumulated proposal pool is periodically reviewed to update the experiential memory. Its role is to prevent unstable memory changes caused by occasional behaviors or noisy reflections, which is essential for building reliable time-evolving experience from long historical sequences. The prompt asks the LLM to group similar proposals and adopt only viewpoints supported by multiple proposals, while weakly supported proposals remain in the proposal pool for further observation. It also applies quality checks before incorporation, ensuring that the updated pattern experience and deficiency experience remain usable, distinguishable, and generalizable. The prompt is shown in Figure~\ref{fig:prompt3}.

\section{Prompt for Groundedness Review}
\label{app:Prompt4}
\begin{figure*}[t]
\centering
 \includegraphics[width=\linewidth]{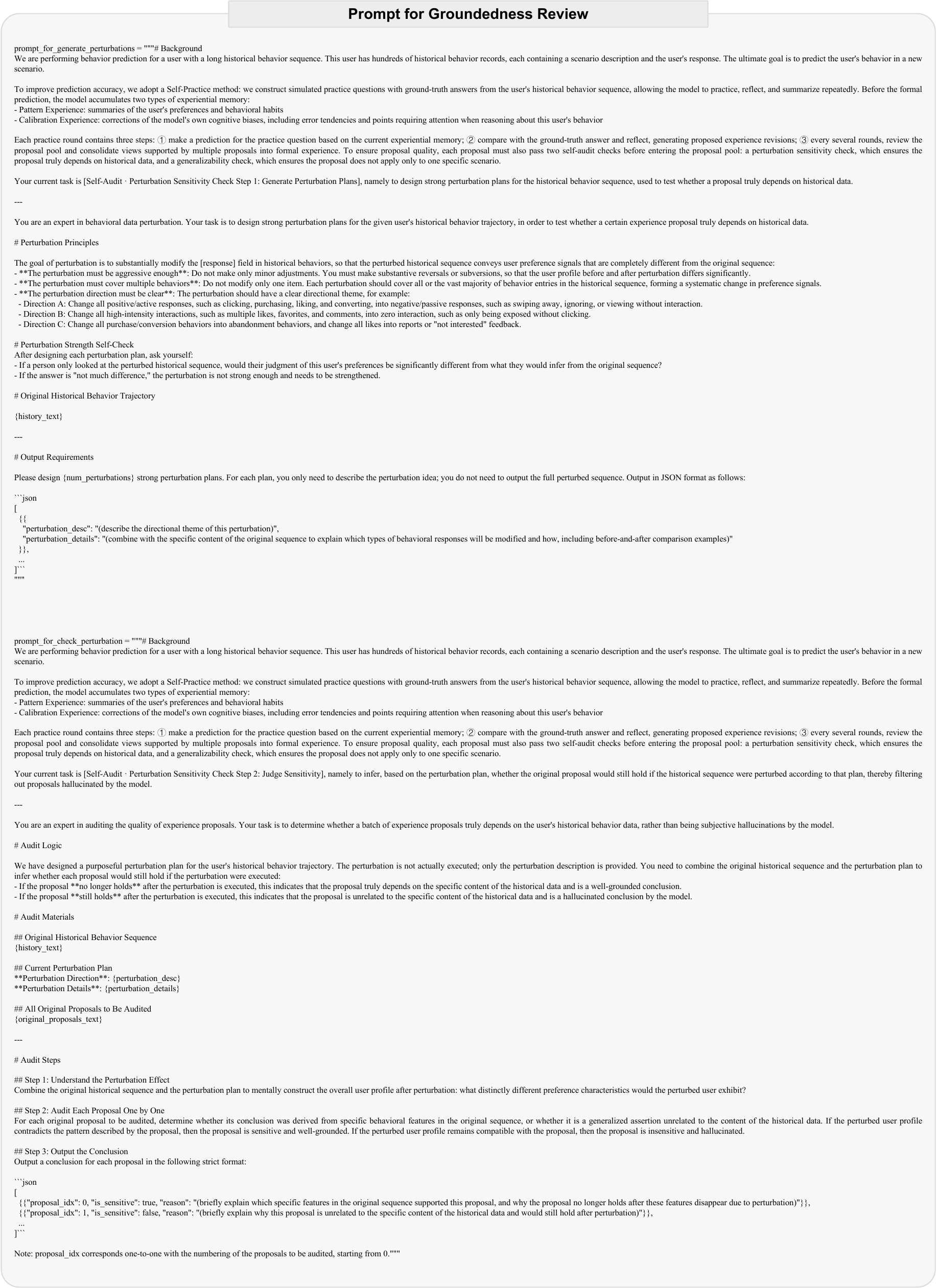} 
\caption{Prompt for groundedness review.}
\label{fig:prompt4}
\end{figure*}
This prompt is used in the self-review mechanism to conduct the perturbation-based groundedness check for reflective proposals. In PraMem, a reliable proposal should be genuinely derived from the historical sequence rather than being a generic or hallucinated statement produced by the LLM. To test this, the prompt first asks the LLM to design strong perturbation plans that substantially change the preference signals conveyed by the historical behavior sequence. Then, based on these perturbation plans, the LLM judges whether each original proposal would still hold after the historical sequence is perturbed; proposals that remain valid under strong perturbation are regarded as insufficiently grounded and are filtered out. The prompt is shown in Figure~\ref{fig:prompt4}.

\section{Prompt for Generalizability Review}
\label{app:Prompt5}
\begin{figure*}[t]
\centering
 \includegraphics[width=\linewidth]{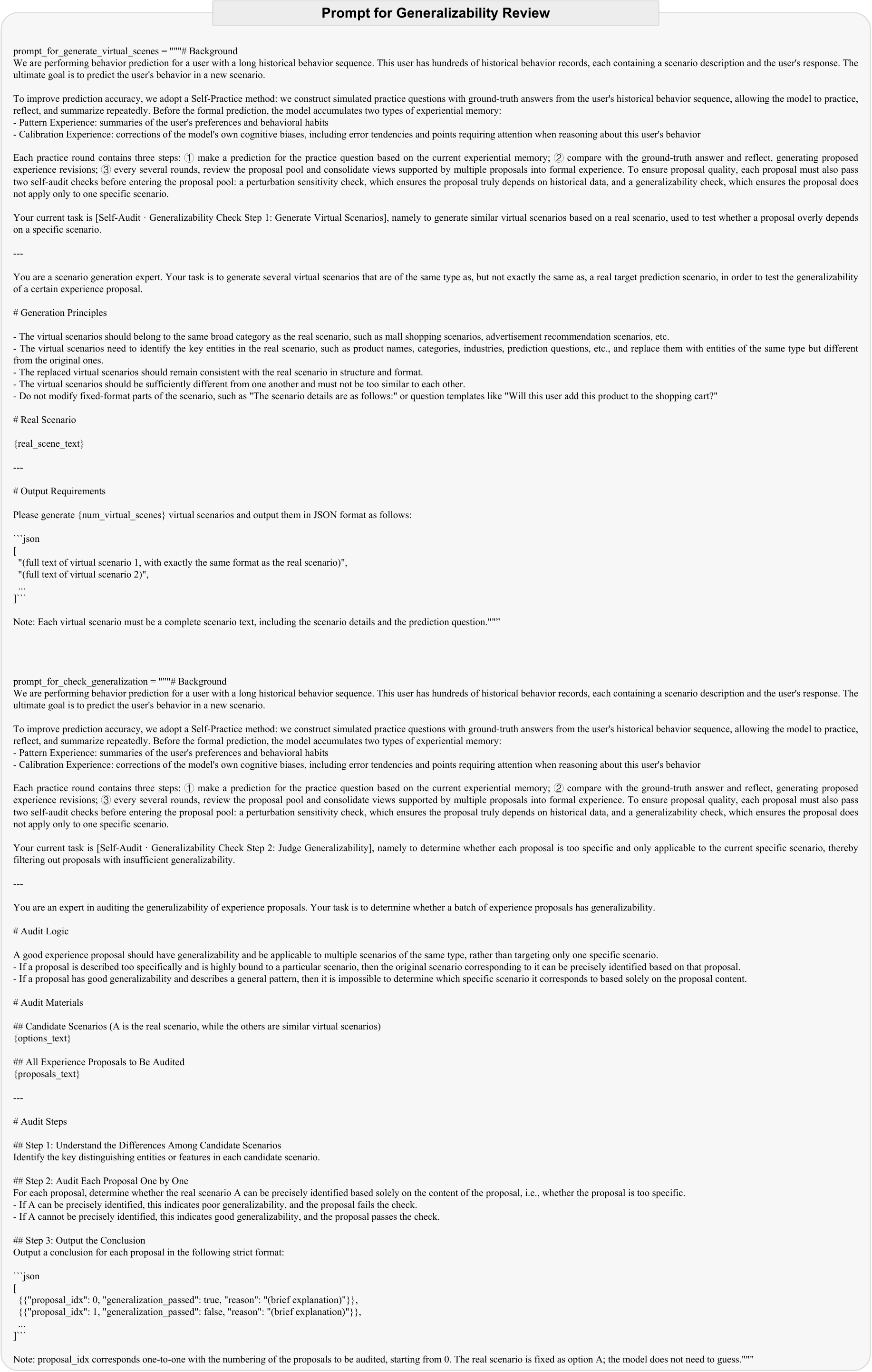} 
\caption{Prompt for generalizability review.}
\label{fig:prompt5}
\end{figure*}
This prompt is used in the self-review mechanism to conduct the generalizability check for reflective proposals. In PraMem, a useful proposal should not merely memorize one practice sample, but should capture experience that can transfer to similar prediction scenarios. To test this, the prompt first generates several virtual scenes that share the same broad type and format as the real prediction scene but differ in key entities or details. Then, the LLM judges whether each proposal can uniquely identify the original real scene among these candidate scenes; if it can, the proposal is considered overly specific to the current practice sample and is filtered out. The prompt is shown in Figure~\ref{fig:prompt5}.

\section{Case Study}
\label{app:case}

\begin{figure*}[t]
\centering
 \includegraphics[width=\linewidth]{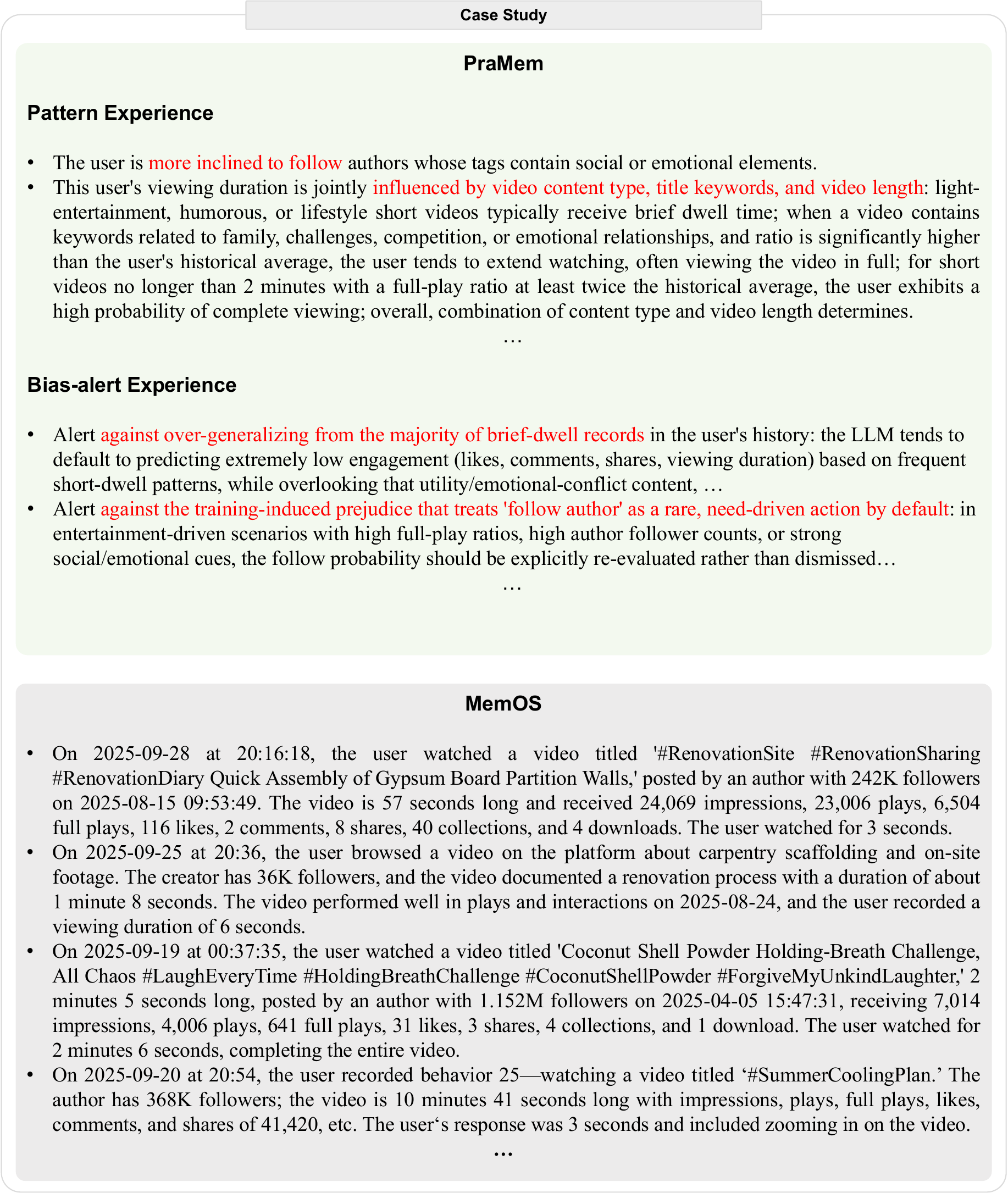} 
\caption{A case study showing that PraMem distills abstract behavioral patterns and bias alerts through practice, whereas the baseline MemOS merely records factual historical behaviors.}
\label{fig:case}
\vspace{-10pt}  
\end{figure*}

To more intuitively demonstrate the essential advantages of the experiential memory constructed by PraMem over traditional memory management methods, we conduct a case study that compares the memory content maintained for the same user under PraMem and the representative baseline method MemOS, as shown in Figure~\ref{fig:case}.

In this case, the memory distilled by PraMem consists of two parts: on one hand, the pattern experience characterizes the user's behavioral regularities in a highly condensed manner, such as "the user is more inclined to follow authors whose tags contain social or emotional elements" and "the viewing duration is jointly determined by video content type, title keywords, and video length"; on the other hand, the bias-alert experience explicitly flags the cognitive biases that the LLM tends to fall into when facing this user, such as "alert against over-generalizing from frequent brief-dwell records in the history into extremely low engagement predictions" and "alert against the training-induced prejudice that treats following an author as a rare, need-driven action by default". 
In contrast, the memory maintained by MemOS is primarily a factual recount of the user's historical behaviors, recording concrete entries such as "on 2025-09-28 at 20:16:18, the user watched a 57-second video about gypsum board partition walls for 3 seconds", with each entry confined to an objective description of a single behavior.

From this, it can be seen that the memories constructed by PraMem and the baseline method differ fundamentally in nature: the baseline's memory is a record of facts, and even after compression and retrieval, its content still remains at the level of "what has happened", requiring the LLM to induce behavioral patterns on the fly from fragmented records during the prediction stage, while being powerless against the model's intrinsic cognitive biases; in contrast, PraMem's memory is the experience distilled through iterative practice, which not only explicitly presents the user's latent behavioral patterns in a high-level abstract manner, but also pre-alerts the LLM against the systematic errors it tends to commit on this user, thereby directly acting on the prediction process and genuinely resolving the two core challenges of long-horizon behavior prediction.

\section{Efficiency Report}
\label{app:time}
\begin{table}[]
\centering
\begin{tabular}{lcc}
\toprule
         & Construction Time (min) & Overall F1 \\
\midrule
Mem0~\citep{chhikara2025mem0}     & 34.1                    & 26.4       \\
MemOS~\citep{li2025memos}    & 30.6                    & 26.2       \\
ProEX~\citep{zhang2026proex}    & 24.7                    & 26.9       \\
\midrule
\textbf{PraMem (Ours)} & 55.8                    & 31.6       \\
\bottomrule
\end{tabular}
\vspace{5pt}
\caption{Efficiency report of PraMem and baselines.}
\label{table:eff}
\vspace{-10pt}  
\end{table}
We report in Table~\ref{table:eff} the average per-user time cost of memory construction for PraMem and three advanced memory-based methods. Since PraMem performs extensive iterative practice over the historical sequence, it naturally incurs higher construction overhead than conventional compression-based methods, while the overall cost remains within an acceptable range.
Moreover, in practical applications, memory construction can be conducted offline, and thus does not increase the latency of online prediction. In this sense, PraMem trades a moderately larger offline construction cost for substantial performance gains, which we believe is acceptable. We further discuss the efficiency issue in detail in the above Limitations section.

\section{Potential Negative Societal Impact}
\label{app:social}
While PraMem is primarily a methodological contribution aimed at improving long-horizon behavior prediction, we acknowledge several potential societal concerns associated with its deployment. First, PraMem relies on lengthy user behavior sequences to build experiential memory, which may involve sensitive personal information such as preferences and interaction histories; if deployed without proper data protection mechanisms, there is a risk of unintended exposure of user behavioral traces, and we therefore encourage practitioners to apply standard privacy-preserving techniques such as anonymization and access control when applying PraMem in real-world scenarios. Second, the improved predictive accuracy of PraMem could potentially be misused for fine-grained user profiling or manipulative targeting in commercial or political contexts; although our work focuses on benign applications such as recommendation and user simulation, we advocate for responsible use of behavior prediction technologies and compliance with relevant ethical guidelines and regulations.



\end{document}